\documentclass[letterpaper, 10 pt, conference]{ieeeconf}  

\IEEEoverridecommandlockouts                              

\usepackage{epsfig} 

\usepackage{makecell}
\usepackage{subcaption}
\usepackage{xcolor}
\usepackage{ragged2e}

\usepackage{graphicx}
\usepackage{amsmath}
\usepackage{adjustbox}
\usepackage{amssymb}
\usepackage{booktabs}
\usepackage{bm}
\usepackage{multirow}
\title{\LARGE \bf
Socially-Aware Robot Navigation Enhanced by Bidirectional Natural Language Conversations Using Large Language Models
}

\author{Congcong Wen*, Yifan Liu*, Geeta Chandra Raju Bethala, Shuaihang Yuan, Hao Huang,\\ Yu Hao, Mengyu Wang, Yu-Shen Liu,  Anthony Tzes, and Yi Fang
\thanks{Congcong Wen, Geeta Chandra Raju Bethala, Shuaihang Yuan, Hao Huang, Yu Hao, and Yi Fang are with Embodied AI and Robotics (AIR) Lab, New York University, New York, USA and NYUAD Center for Artificial Intelligence and Robotics, New York University Abu Dhabi, Abu Dhabi, UAE.
{\tt\small \{cw3437, gb2643, sy2366, hh1811, yh3252, yf23\}@nyu.edu}}
\thanks{Yifan Liu is with the UCLA Mobility Lab and Mobility Center
of Excellence, Los Angeles, Los Angeles, USA. {\tt\small bmmliu@ucla.edu }}
\thanks{Mengyu Wang is with the Harvard AI and Robotics Lab, Harvard University, Boston, USA. {\tt\small mengyu\_wang@meei.harvard.edu}}%
\thanks{Yu-Shen Liu is with the School of Software, Tsinghua University, Beijing, P. R. China. {\tt\small liuyushen@tsinghua.edu.cn}}%
\thanks{Anthony Tzes is with the NYUAD Center for Artificial Intelligence and Robotics, New York University Abu Dhabi, Abu Dhabi, UAE. {\tt\small anthony.tzes@nyu.edu}}%
\thanks{* indicates equal contribution.}
}

\begin{document}

\maketitle
\thispagestyle{empty}
\pagestyle{empty}

\begin{abstract}

Robot navigation is crucial across various domains, yet traditional methods focus on efficiency and obstacle avoidance, often overlooking human behavior in shared spaces. With the rise of service robots, socially aware navigation has gained prominence. However, existing approaches primarily predict pedestrian movements or issue alerts, lacking true human-robot interaction. We introduce Hybrid Soft Actor-Critic with Large Language Model (HSAC-LLM), a novel framework for socially aware navigation. By integrating deep reinforcement learning with large language models, HSAC-LLM enables bidirectional natural language interactions, predicting both continuous and discrete navigation actions. When potential collisions arise, the robot proactively communicates with pedestrians to determine avoidance strategies. Experiments in 2D simulation, Gazebo, and real-world environments demonstrate that HSAC-LLM outperforms state-of-the-art DRL methods in interaction, navigation, and obstacle avoidance. This paradigm advances effective human-robot interactions in dynamic settings. Videos are available at https://hsacllm.github.io/.
\end{abstract}

\section{INTRODUCTION}

Robot navigation is essential for enabling movement while avoiding collisions across applications such as autonomous vehicles~\cite{zhao2024triplemixer}, drones~\cite{10.1109/ICRA48506.2021.9561188}, and robots~\cite{wen2025zero,wen2024secure}. However, traditional approaches prioritize efficiency and obstacle avoidance while neglecting social cues and human intentions, crucial for coexistence in shared spaces.

With advances in robotics, socially-aware navigation~\cite{kruse2013human} has emerged as a key area, equipping robots with the ability to navigate while respecting social norms and engaging in natural interactions. Beyond safe movement, robots must predict human behavior and adhere to social etiquette. For instance, airport navigation robots leverage language comprehension and social awareness to interact naturally with passengers from diverse backgrounds. Earlier methods focused on human motion prediction for safer path planning, using models like Social Force Model (SFM)\cite{helbing1995social}, Social LSTM\cite{7780479}, and Social GAN~\cite{gupta2018social}. Gesture and gaze recognition techniques~\cite{keskin2012hand} further improved predictions but often led to unnecessary detours or collisions. Recent approaches introduce pedestrian alerts through "Beeping," "Voice Noticing," and "Nudging"~\cite{9341519, ma2022reinforcement, dugas2020ian}. However, these one-way signals can be ambiguous, ineffective in noisy environments, and intrusive, diminishing user experience and robot acceptance.

In this paper, we propose the Hybrid Soft Actor-Critic with Large Language Model (HSAC-LLM), integrating deep reinforcement learning with large language models for socially-aware robot navigation with natural language capabilities. Fig.\ref{fig:overview} compares our approach with existing methods: non-interaction strategies (Detour\cite{yao2021crowd}) and unidirectional voice interaction (Beep~\cite{9341519}). Our method enables robots to better understand human intentions in complex environments. We process environmental data through PreNet networks to capture the current state, including voice input from conversations initiated when potential collisions are detected. Our HSAC-LLM model simultaneously predicts continuous actions (linear/angular speeds) and discrete actions (interaction codes), while generating appropriate voice notifications via the LLM. Experiments in 2D simulations, Gazebo environments, and real-world settings demonstrate HSAC-LLM's superior performance over state-of-the-art DRL algorithms. This approach enhances navigation safety and efficiency while fostering harmonious human-robot coexistence in social scenarios. The main contributions are as follows:

\begin{figure}
  \centering
  \medskip
  \includegraphics[width=0.95\linewidth]{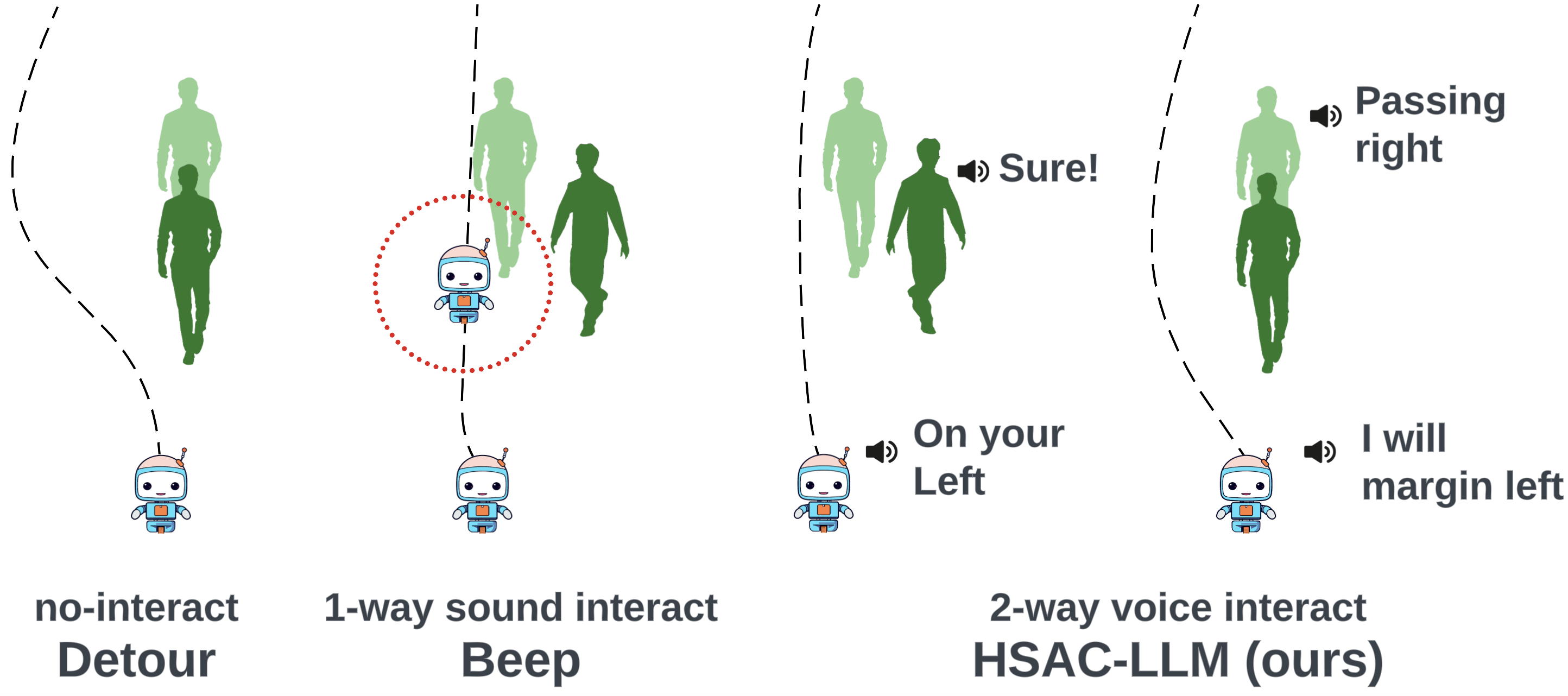}
  \caption{Strategies used to avoid incoming collision with pedestrians. Left: Detour for a longer route to prevent collision~\cite{yao2021crowd}. Middle: Beeping to alert pedestrians and create space for the robot~\cite{9341519}. Right: Interactive voice interaction, enabling natural language conversations between robots and pedestrians (our approach).
  }
  \label{fig:overview}
\end{figure}




\vspace{-0.08cm}
\begin{itemize}
    \item We explore the integration of socially-aware navigation and natural language interaction, addressing the challenge of enabling robots to navigate complex and dynamic environments while respecting social norms.

    \item We develop HSAC-LLM, a unified framework combining deep reinforcement learning and large language models that empower robots to understand and respond to human intentions through natural conversation.

    \item We introduce comprehensive evaluation metrics for socially-aware navigation and rigorously assess its performance from three aspects: safety, efficiency, and human comfort.

    \item We validate our approach through experiments in simulated and real-world environments, demonstrating superior performance over existing methods.
\end{itemize}

\section{RELATED WORK}

\subsection{Robot Navigation}

Safely reaching a destination while avoiding obstacles has long been a key challenge in mobile robotics. Early approaches relied on deterministic algorithms such as Simultaneous Localization and Mapping (SLAM) \cite{1638022}, which generated precise environment maps using sensor data like LiDAR and vision-based inputs \cite{temeltas2008slam}. Path-planning methods, including A* and Dijkstra’s, were then applied to compute collision-free trajectories \cite{hart1968formal}. However, these approaches struggled in dynamic environments due to high computational demands and limited adaptability \cite{balasuriya2016outdoor}. The emergence of Deep Reinforcement Learning (DRL) introduced a paradigm shift by leveraging raw sensor data to learn navigation policies without explicit mapping. DRL-enabled mapless navigation has demonstrated adaptability in dynamic environments with moving obstacles \cite{banino2018vector}. Additionally, combining DRL with convolutional neural networks (CNNs) enhances feature extraction from sensor inputs, improving navigation in complex scenarios \cite{tai2017virtual}. Virtual-to-real transfer techniques further optimize DRL training, reducing real-world deployment costs \cite{zhang2019vr}.

\begin{figure*}[t]
    \centering
    \medskip
    \includegraphics[width=1.0\linewidth]{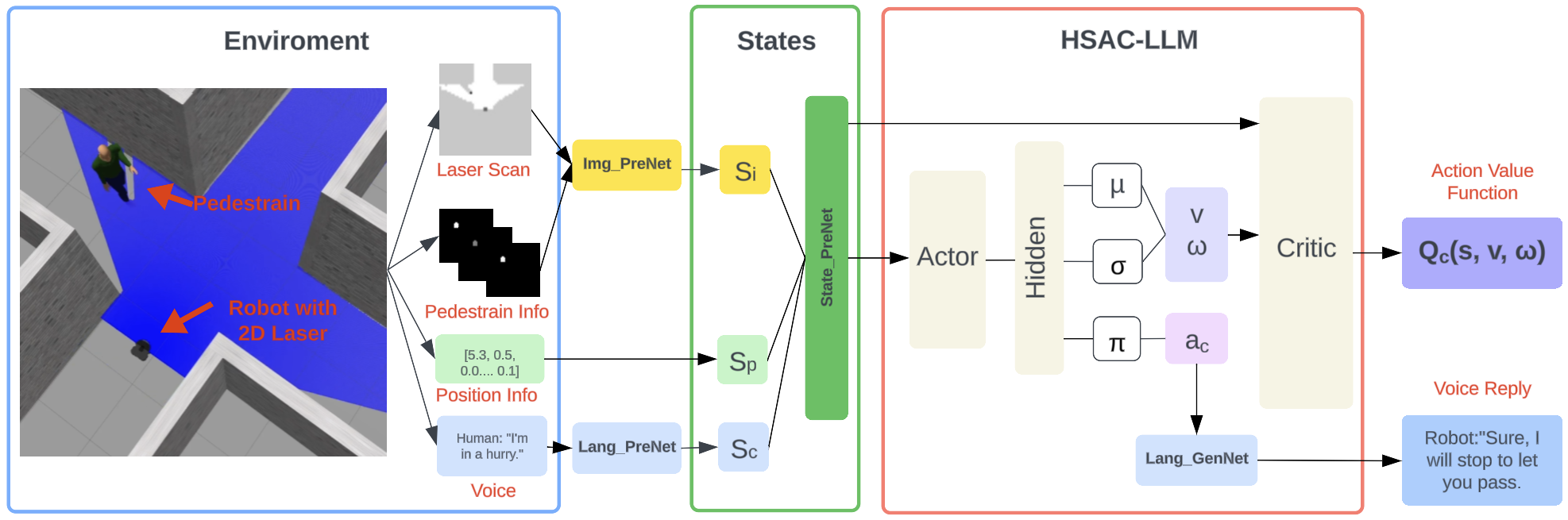}
    \caption{Illustration of a typical environment and the architecture of the proposed HSAC-LLM model.
    }
    \label{fig:overall_architecture}
\end{figure*}

\subsection{Human-Robot Interaction in Robot Navigation}

Robots increasingly interact with humans in roles like food delivery~\cite{chen2021adoption}, tour guiding~\cite{del2019lindsey}, and healthcare~\cite{reis2020service}. Ensuring comfort, naturalness, and sociability is key for human-aware navigation~\cite{KRUSE20131726}. Early research focused on predicting human movements for safer path planning, employing models like the Social Force Model (SFM)\cite{helbing1995social}, Social LSTM\cite{7780479}, Social GAN~\cite{gupta2018social}, and MSA3C~\cite{he2022multirobot}. Computer vision techniques further enabled gesture recognition~\cite{keskin2012hand}, but often prioritized navigation over interaction, leading to detours or immobilization. Recent methods integrate proactive robot-human interactions. Techniques like "beeping" effectively alert pedestrians, improving success and reducing collisions in indoor navigation~\cite{9341519} and elevator entry tasks~\cite{ma2022reinforcement}. Gestures like "nudging" also enhance navigation performance~\cite{dugas2020ian}. However, these methods rely on the robot’s initiative, and assertive actions may not always ensure a comfortable, natural interaction experience.

\subsection{Discrete-Continuous Hybrid Action Space}

Managing a Discrete-Continuous hybrid action space is crucial for enabling robots to control continuous velocity while engaging in discrete human interactions. While DRL algorithms for hybrid action spaces have been validated in robotic arm tasks~\cite{neunert2020continuousdiscrete}, adapting them to robot navigation remains challenging. Prior approaches have addressed this by discretizing velocity actions~\cite{9341519} or using Reinforcement Learning for interaction selection~\cite{9811662}, but the latter often relies on extensive human-labeled data, introducing subjectivity. Hybrid action spaces are also prevalent in video games, where players control movements continuously while making discrete decisions like jumping or braking. DRL methods such as Multi-Pass deep Q-Networks~\cite{bester2019multipass} and Hybrid Soft Actor-Critic~\cite{delalleau2019discrete} have demonstrated success in Gym environments, providing a foundation for our approach to human-robot interaction in navigation tasks.

\section{Methods}


\subsection{Overview}

We introduce the Hybrid Soft Actor-Critic with Large Language Model (HSAC-LLM), a novel paradigm that combines deep reinforcement learning (DRL) with Large Language Models (LLMs) to facilitate bidirectional natural language interactions between robots and pedestrians during navigation. This approach ultimately enables intelligent obstacle avoidance while providing a comfortable and interactive experience. Fig.~\ref{fig:overall_architecture} shows a typical environment and the architecture of the proposed HSAC-LLM model. The radar scans and pedestrian information are first input to an Img\_PreNet network, obtaining the state $\bm{S}_i$. Concurrently, the robot's position information is captured as state $\bm{S}_p$. When potential collision risks are detected, the robot either initiates or responds to vocal interactions with humans. These interactions are processed through a Lang\_PreNet network to generate the conditional state $\bm{S}_c$. Combining these states, the robot derives its subsequent actions and velocities by feeding the aggregated states into the proposed HSAC-LLM. Crucially, this model leverages both state and condition in its decision-making, represented as $\pi(\bm{a} | [\bm{S}_i, \bm{S}_p], \bm{S}_c)$. Once an appropriate action is determined, the robot leverages Lang\_GenNet to provide vocal feedback to pedestrians, facilitating real-time communication and enhancing safety during navigation.

\subsection{PreNet}

Fig.~\ref{fig:overall_architecture} illustrates our input data sources: a 180-degree 2D Laser scan grid map representing the robot's spatial configuration, a three-channel pedestrian grid map indicating the nearby pedestrian's location and speed, a 9-dimensional relative position vector for target position and orientation over three time steps, and voice messages from pedestrians. Rather than directly integrating these data into HSAC-LLM, we employ three preprocessing networks to convert image and voice data from the environment into state vectors.
\paragraph{Img\_PreNet}
Following~\cite{yao2021crowd}, we initiate feature extraction using three convolutional layers and three max pooling layers on laser scan data and pedestrian information grid map, producing a 512-dimensional vector $\bm{S}_i$ through a fully connected layer.

\paragraph{Lang\_PreNet}
When the robot perceives a potential collision with a human, it engages in a vocal conversation, either initiating or responding. To better understand these voice information, we fed these messages into a Lang\_PreNet network to produce state vector $\bm{S}_c$ for the robot policy condition.

\paragraph{State\_PreNet}
The image info vector $\bm{S}_i$,  the relative position info vector $\bm{S}_p$ and condition code vector $\bm{S}_c$ are concatenated and passed into the State\_PreNet which contains one fully connected layer with 512 units to form a 512-dimensional vector $S$, which will serve as the environment state for the HSAC-LLM model.

\subsection{HSAC-LLM}
In this section, we introduce HSAC-LLM, a model that enables bidirectional communication between robots and pedestrians to coordinate trajectories and prevent collisions. It comprises two main components: the \textit{HSAC} (Hybrid Soft Actor Critic) module and the \textit{Lang\_GenNet} module. HSAC models both continuous and discrete robot actions, while Lang\_GenNet serves a complementary purpose, which interprets and conveys the discrete outputs as human-friendly information through vocal communication, ensuring a seamless and safe co-navigation experience.

\paragraph{HSAC Module}
During robot navigation, behavior involves both continuous and discrete elements. We categorize the robot's motion controls, angular and linear velocities, as continuous actions, while post-interaction decisions are discrete actions. The standard SAC (Soft Actor-Critic) approach was originally introduced to model continuous actions for robotic tasks. Inspired by ~\cite{delalleau2019discrete}, we propose a Hybrid SAC model tailored specifically for our current navigation task, emphasizing human-robot interactions. Fig.~\ref{fig:overall_architecture} shows the architecture of the proposed HSAC-LLM model. Specifically, we first utilize the actor to generate a shared hidden state representation $\bm{h}$ that is used to produce both a discrete distribution $\bm{\pi}$ as well as the mean $\bm{\mu}$ and standard deviation $\bm{\sigma}$ of the continuous component. The discrete action $\bm{a}_c$ is sampled from $\bm{\pi}$ and while the continuous actions, including  linear velocity $\bm{v}$ and angular velocity $\bm{\omega}$, are computed as according the following formulation:
\begin{equation}
    [\bm{v}, \bm{\omega}] = tanh(\bm{\mu}+\epsilon \bm{\sigma})
\end{equation}
After obtaining the continuous and discrete actions, the critic network takes both the state $\bm{s}$ and the continuous action $\bm{\mu}$ and $\bm{\omega}$ as input and predicts the Q-values of all discrete actions in its output layer. And the HSAC's policy update formula will be:
\begin{equation}
    \pi^* = \arg \min_\pi \mathbb{E}_{\bm{a} \sim \pi} [ \alpha \log \pi(\bm{a}|[\bm{S}_i, \bm{S}_p], \bm{S}_c)) - Q(\bm{s}, \bm{\mu}, \bm{\omega}, \bm{a}_c) ]
\end{equation}

\paragraph{Lang\_GenNet Module}
The Lang\_GenNet module, predicated on the architecture of Large Language Models (LLMs), is meticulously crafted to transform the action vector $\bm{a}_c$, derived from the HSAC model's output, into coherent natural language interactions with pedestrians. In this study, the module is instantiated leveraging the capabilities of the GPT-4 model, showcasing a sophisticated integration of state-of-the-art language understanding and generation techniques.

\subsection{Reward Function}

We define the reward for the robot at timestep $t$ as $r^t$, which is composed of four components: $r^t_{goal}$, $r^t_{shaping}$, $r^t_{step}$, and $r^t_{interact}$. For $r^t_{goal}$, the robot is awarded a positive reward $r_{\text{arrive}}$, if it successfully reaches its destination without its minimum distance to pedestrian $d^t_{min}$ falling below the safety threshold $d_{safe}$ at any point during its journey. However, if the robot reaches its goal but at any moment the minimum distance to a pedestrian $d^t_{min}$ drops below the safety threshold $d_{safe}$, a negative penalty proportional to the infraction is applied to $r_{\text{arrive}}$. In the event of a collision, $r^t_{goal}$ incurs a negative penalty, denoted as $r_{\text{collide}}$. In all other cases, $r^t_{goal}$ is set to 0. We formalize this as follows:
\begin{equation}\label{eq:reward_goal}
    r^t_{goal} ={} 
  \begin{cases}
    r_{arrive} & \quad \text{if } arrive,\\
    r_{arrive}-\epsilon_1 (d_{safe} - d^t_{min}) & \quad \text{if }d^t_{min} < d_{safe},\\
    r_{collide} & \quad \text{if } collision,\\
    0 & \quad otherwise,\
  \end{cases}
\end{equation}
where $\epsilon_1$ is a hyper-parameter. And $r^t_{\text{shaping}}$, motivates the robot to advance towards its target position by leveraging its current position $p^t$ and the target position $p_g$. This is defined as follows:
\begin{equation}\label{eq:reward_shaping}
r^t_{shaping} ={} \epsilon_2(||p^{t-1} - p_g|| - ||p^{t} - p_g||) 
\end{equation}
where $\epsilon_2$ is a hyper-parameter. Moreover, we employ the number of steps taken as a negative penalty, $r^t_{step}$, which encourages the robot to select shorter paths, thereby increasing its movement efficiency.

In addition, we introduced an interaction reward, $r^t_{interact}$, designed to encourage proper interaction between the robot and humans. When the robot initiates a dialogue request, $R_r$, and this request is in alignment with the current pedestrian trajectory, a reward of $r^t_{req}$ is granted. Conversely, if it is the pedestrian who initiates the dialogue request, $R_p$, and the robot's path planning generated by the model aligns with the pedestrian's request, a reward of $r^t_{res}$ is bestowed. In all other instances, $r^t_{interact}$ is set to 0. The formulation is as follows:
\begin{equation}\label{eq:reward_interact}
    r^t_{interact} ={}
      \begin{cases}
        r^t_{req} & \quad \text{if }R_r \text{ is not none}\\
        r^t_{res} & \quad \text{if }R_p \text{ is not none}\\
        0 & \quad otherwise.\\
      \end{cases}
\end{equation}
Hence, the final reward $r^t$ can be calculated as:
\begin{equation}\label{eq:reward_all}
    r^t =r^t_{goal} + r^t_{shaping} + r^t_{step} + r^t_{interact}
\end{equation}

\section{Experiments}
Our algorithm is trained within a 2D simulator, utilizing the OpenCV framework as detailed in~\cite{chen2020distributed} and~\cite{yao2021crowd}. Subsequently, we evaluate its performance across three distinct environments: the 2D simulation environment, the Gazebo simulation environment, and the real-world environment.

\begin{figure*}[t]
    \centering
    \medskip
    \includegraphics[width=0.80\linewidth]{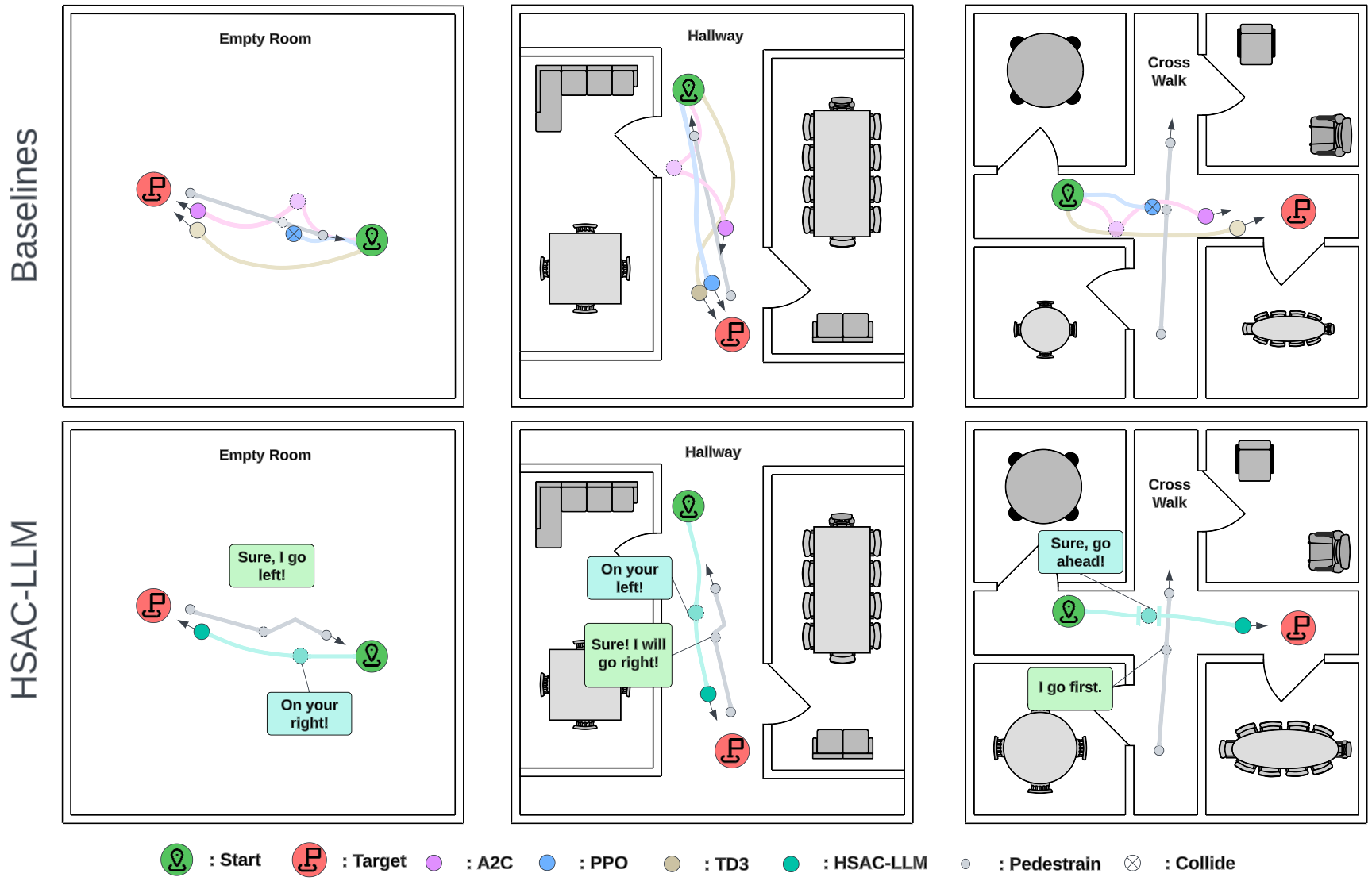}
    \caption{Visualization of navigation trajectories in the Square, Hallway, and Crosswalk scenarios for baseline models (top) and our HSAC-LLM model (bottom).
    }
    \label{fig:results_simu}
\end{figure*}

\subsection{Experiments in 2D simulator}
\subsubsection{Experimental Setting} In the 2D simulator, all pedestrians were controlled by a customized version of the Reciprocal Velocity Obstacles (RVO) algorithm~\cite{4543489}, which enabled them to respond to the robot's requests to dodge by performing stop and margin actions. We refer to this augmented model as Interactive Reciprocal Velocity Obstacles (IRVO). Our experimental focus was on scenarios involving a single pedestrian, closely mirroring real-world conditions for voice-based human-robot interactions. For all algorithms, the action spaces for linear and angular velocities were set to $v\in [0, 0.6]$ m/s and $\omega\in [-0.9, 0.9]$ rad/s, respectively; the velocity of the pedestrian was fixed at 0.5 m/s. Equation~\ref{eq:reward_all} served as the reward function for the baseline methods. The GPT-4 model from OpenAI was selected as the LLM for the HSAC-LLM model. We employed a four-dimensional discrete action space for action codes, denoted as $\bm{a}_c \in$ {none, stop, margin-right, margin-left}, where ``none" indicates no interaction, ``stop" signals the pedestrian to halt, and ``margin" commands the pedestrian to move to the left or right for creating space.

Considering the complexity of real-world navigation environments, we designed three distinct scenes within the simulator: \textit{Square}, \textit{Hallway}, and \textit{Crosswalk}, as illustrated in Fig.~\ref{fig:results_simu}. To simulate varied navigation challenges, the starting and ending positions of the robot and pedestrian within each scene adhere to specific patterns. In the \textit{Square} scene, the robot and pedestrian are placed at random locations. For the \textit{Hallway} scene, both agents start at opposite ends, moving toward each other. In the \textit{Crosswalk} scenario, agents start at different corridor ends, crossing paths. The primary goal of the robot is to navigate to its destination, prioritizing pedestrian safety and collision avoidance. The parameters of the reward function were set as follows: $r_{arrive} = 500$, $r_{collide} = -500$, $r_{step} = -5$, $d_{safe} = 1.0$, $\epsilon_1 = -50$ and $\epsilon_2 = 200$.

\subsubsection{Metrics} Following \cite{francis2023principles}, we propose the following metrics to evaluate navigation performance:
\begin{itemize}
\item \textbf{Success}: Ratio of safe target arrivals to total attempts.
\item \textbf{Collision}: Ratio of collisions to total attempts.
\item \textbf{Step}: Average steps for successful navigation.
\item \textbf{Training Time}: Model training duration.
\item \textbf{Human Collisions}: Ratio of human collisions to total attempts.
\item \textbf{Minimum Distance to Human}: Shortest robot-human distance during successful navigation.
\item \textbf{Comfort Space Compliance}: Ratio of instances maintaining minimum comfort distance.
\end{itemize}

\subsubsection{Baselines}
We incorporated state-of-the-art DRL algorithms tailored for continuous action spaces as our baseline methods, including Advantage Actor Critic (A2C)~\cite{mnih2016asynchronous}, Proximal Policy Optimization (PPO)\cite{schulman2017proximal}, and Twin Delayed Deep Deterministic Policy Gradients (TD3)\cite{fujimoto2018addressing}. 

\subsubsection{Results}
All models were trained on an i7-10750H CPU and an NVIDIA RTX 2070 GPU for 128,000 steps. The results of 500 tests in each scene are presented in Table~\ref{tab:results_simu}, and the trajectories of various agents are depicted in Fig.~\ref{fig:results_simu}. Throughout testing, a reverse-request probability of $p_r = 0.5$ was employed, ensuring an equal chance for both the robot and the pedestrian to initiate an avoidance request.

\begin{table*}[] 
\caption{Performance of our HSAC-LLM model compared with different baseline models in three scenarios. MDH stands for Minimum Distance to Human, and CSC denotes Comfort Space Compliance}
\label{tab:results_simu}
\centering
\resizebox{0.8\textwidth}{!}{
\begin{tabular}{ccccccccc}
\toprule
\textbf{Scene}             & \textbf{Method} & \textbf{Success}$\uparrow$ & \textbf{Collision}$\downarrow$ & \textbf{Step}$\downarrow$ & \textbf{Training Time}$\downarrow$ & \textbf{Human Collisions}$\downarrow$ & \textbf{MDH}$\uparrow$ & \textbf{CSC}$\uparrow$\\
\midrule
\multirow{4}{*}{Square}    & A2C             & 0.69 ± 0.045     & 0.28 ± 0.030     & 27.38         & \textbf{0.9}                 & 0.28                        & 0.25                   & 0.81                              \\
                           & TD3             & 0.85 ± 0.009     & 0.14 ± 0.008     & 21.21         & 2.6                 & 0.14                        & 0.28                   & 0.80                               \\
                           & PPO             & 0.58 ± 0.028     & 0.42 ± 0.023     & 37.81         & 8.0                   & 0.39                        & 0.23                   & 0.82                              \\
                           & HSAC-LLM        & \textbf{0.91 ± 0.022}     & \textbf{0.08 ± 0.022}     & \textbf{19.96}         & 1.6                 & \textbf{0.07}                        & \textbf{0.31}                   & \textbf{0.85}                              \\
\midrule
\multirow{4}{*}{Hallway} & A2C & 0.28 ± 0.038 & 0.59 ± 0.096 & 58.85 & \textbf{0.9} & 0.32 & 0.42 & 0.91 \\
& TD3 & 0.77 ± 0.044 & 0.20 ± 0.035 & 41.12 & 5.0 & 0.21 & 0.30 & 0.89 \\
& PPO & 0.63 ± 0.037 & 0.37 ± 0.037 & 30.30 & 6.6 & 0.08 & \textbf{0.50} & 0.91  \\
& HSAC-LLM & \textbf{0.93 ± 0.013} & \textbf{0.07 ± 0.013} & \textbf{27.10} & 1.9 & \textbf{0.06} & 0.32 & \textbf{0.93} \\
\midrule
\multirow{4}{*}{Crosswalk} & A2C & 0.14 ± 0.016 & 0.77 ± 0.029 & 43.83 & \textbf{1.1} & 0.58 & \textbf{1.28} & \textbf{0.99} \\
& TD3 & 0.85 ± 0.053 & 0.14 ± 0.044 & 41.67 & 3.0 & 0.14 & 0.33 & 0.90 \\
& PPO & 0.46 ± 0.065 & 0.53 ± 0.065 & 33.33 & 10.8 & 0.53 & 0.54 & 0.91 \\
& HSAC-LLM & \textbf{0.91 ± 0.022} & \textbf{0.09 ± 0.022} & \textbf{32.80} & 2.0 & \textbf{0.08} & 0.40 & 0.92 \\
\bottomrule
\end{tabular}}
\end{table*}


From Table~\ref{tab:results_simu}, we observe that our HSAC-LLM algorithm outperforms the state-of-the-art DRL algorithms in all three scenarios, achieving the highest success rates, the lowest collision rates, and the shortest step counts. Specifically, in the Square scenario, we achieved a success rate of 0.91, a collision rate of 0.08, a minimum distance to humans of 0.31, and a Comfort Space Compliance of 0.85. The significantly lower collision rate compared to other methods highlights our model's ability to greatly reduce collisions through natural language-based interactions. In the more complex Hallway and Crosswalk scenarios, where collisions between humans and robots are more likely, our model still maintains collision rates of only 0.07 and 0.09, respectively. This is in contrast to other methods, which rely solely on unidirectional detouring to avoid collisions, a tactic that often results in collisions in slightly more complex environments due to insufficient evasion time. However, our HSAC-LLM model flexibly engages in natural language-based communication with pedestrians before potential collisions occur, fully understanding their intentions and eliminating ambiguities in communication, thus effectively preventing any impending collision.

\begin{figure*}[tp]
  \centering
  \medskip
  \includegraphics[width=0.9\linewidth]{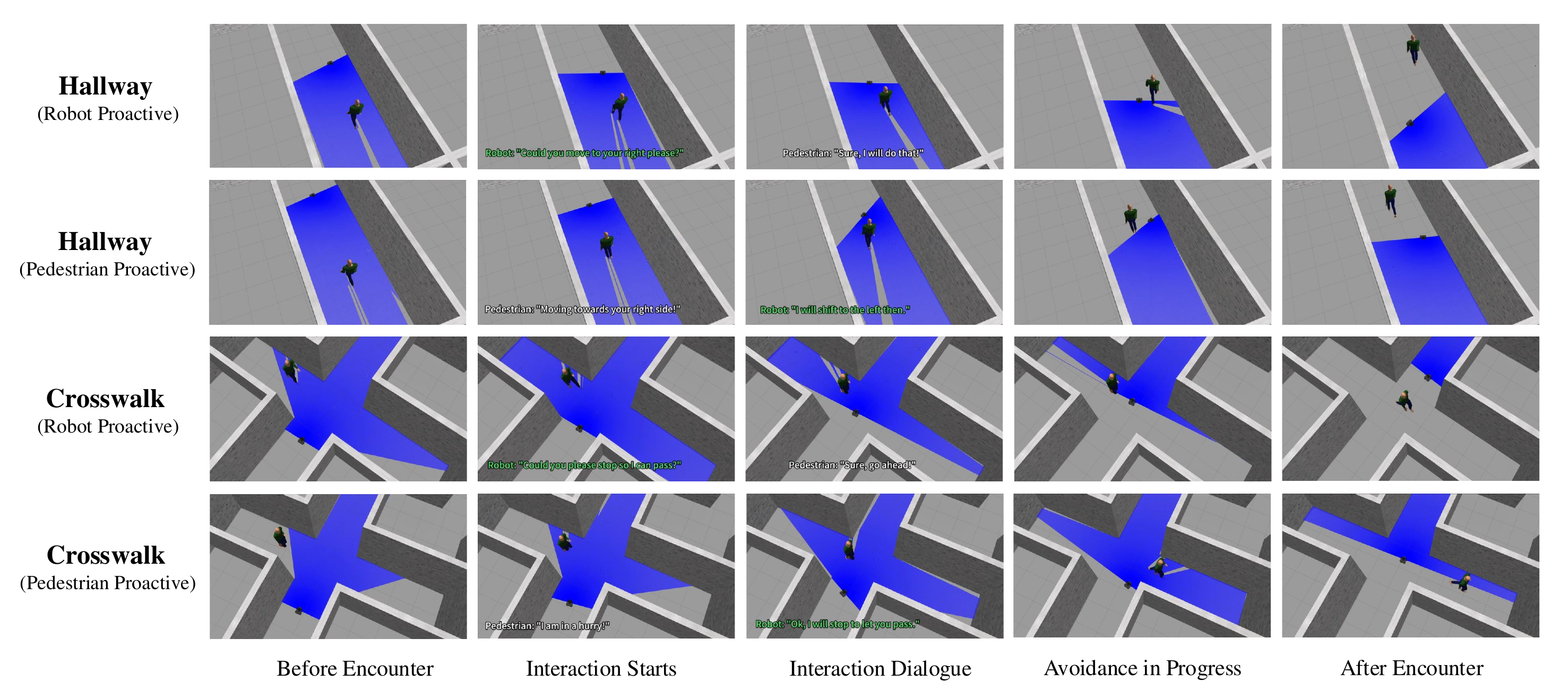}
  \caption{Interaction Scenarios in Hallway and Crosswalk Environments (Robot Proactive and Pedestrian Proactive cases).
  }
  \label{fig:gazebo_env}
\end{figure*}

To more intuitively showcase navigational performance, we randomly selected examples from three scenarios and illustrated the navigation trajectories of both baseline models and our model in Fig.~\ref{fig:results_simu}. We observed that the navigation paths of these baseline models are considerably circuitous, with the PPO model in particular experiencing collisions with pedestrians in both the Square and Crosswalk scenarios. In contrast, our model facilitates bidirectional natural language dialogue between the robot and pedestrians, enabling the dynamic adjustment of passage priorities based on the specific context. Consequently, it efficiently reaches the destination via shorter paths while avoiding collisions.



\subsection{Experiments in Gazebo simulation}

To enhance the real-world applicability of our algorithm, we developed realistic 3D simulation scenarios using the Gazebo simulation environment. Unlike conventional setups where pedestrian speeds are fixed at 0.5 m/s, our approach incorporates a Leg Tracking framework~\cite{7139259} to dynamically estimate pedestrian speeds. This method enables the robot to adaptively determine the optimal distance for initiating or responding to pedestrian interactions, ensuring more flexible and natural movement planning. Importantly, the simulated human in the Gazebo environment is controlled by an actual person, who actively manages speed and direction to replicate realistic human behavior. In parallel, to simulate realistic and naturalistic human-robot interactions, we employed the computer’s microphone and speakers to facilitate pedestrian voice inputs and robotic responses. This setup allows the robot to process spoken commands or requests from pedestrians and generate appropriate verbal replies, creating a lifelike communication exchange.

We conducted comprehensive evaluations across three distinct simulated environments to assess the robot's performance under varying interaction scenarios. The results consistently demonstrate the robot's high responsiveness and adaptability, effectively navigating according to pedestrian preferences. Specifically, the robot behaves proactively to avoid potential conflicts or reactively to accommodate pedestrian-initiated requests, depending on the situation. To intuitively illustrate the robot's navigation performance, Fig.~\ref{fig:gazebo_env}  depicts two primary environments: Hallway and Crosswalk. Each environment contains two types of proactive behaviors: Hallway (Robot Proactive), Hallway (Pedestrian Proactive), Crosswalk (Robot Proactive), and Crosswalk (Pedestrian Proactive). For each scenario, we present five key stages of the interaction process: \textit{Before Encounter}, \textit{Interaction Starts}, \textit{Interaction Dialogue}, \textit{Avoidance in Progress}, and \textit{After Encounter}. These results collectively demonstrate the versatility and robustness of our model in handling diverse interaction contexts, ensuring smooth and efficient navigation while adhering to pedestrian preferences. Notably, our model demonstrates the ability to successfully interpret both explicit commands and implicit cues. Explicit commands, such as “Could you let me pass first?”, are straightforward requests that the robot processes and responds to appropriately. However, the model also handles more subtle implicit cues, such as “I’m in a hurry,” with contextually appropriate behavior. For example, in the Crosswalk (Pedestrian Proactive) scenario, when the pedestrian states, “I’m in a hurry,” the robot responds with: “OK, I will stop to let you pass,” while simultaneously halting to enable the pedestrian to pass safely. This highlights the model's ability to comprehend nuanced human language, extending beyond literal word interpretation to capture the underlying intent. By deeply understanding human preferences and needs, our model effectively plans and executes rational and context-aware navigational routes that align with those needs. This ability to process natural language, interpret human intent, and dynamically adjust navigation strategies underscores the robustness and practicality of our approach for real-world human-robot interactions.

\begin{figure*}[t]
    \centering
    \medskip
    \includegraphics[width=0.9\linewidth]{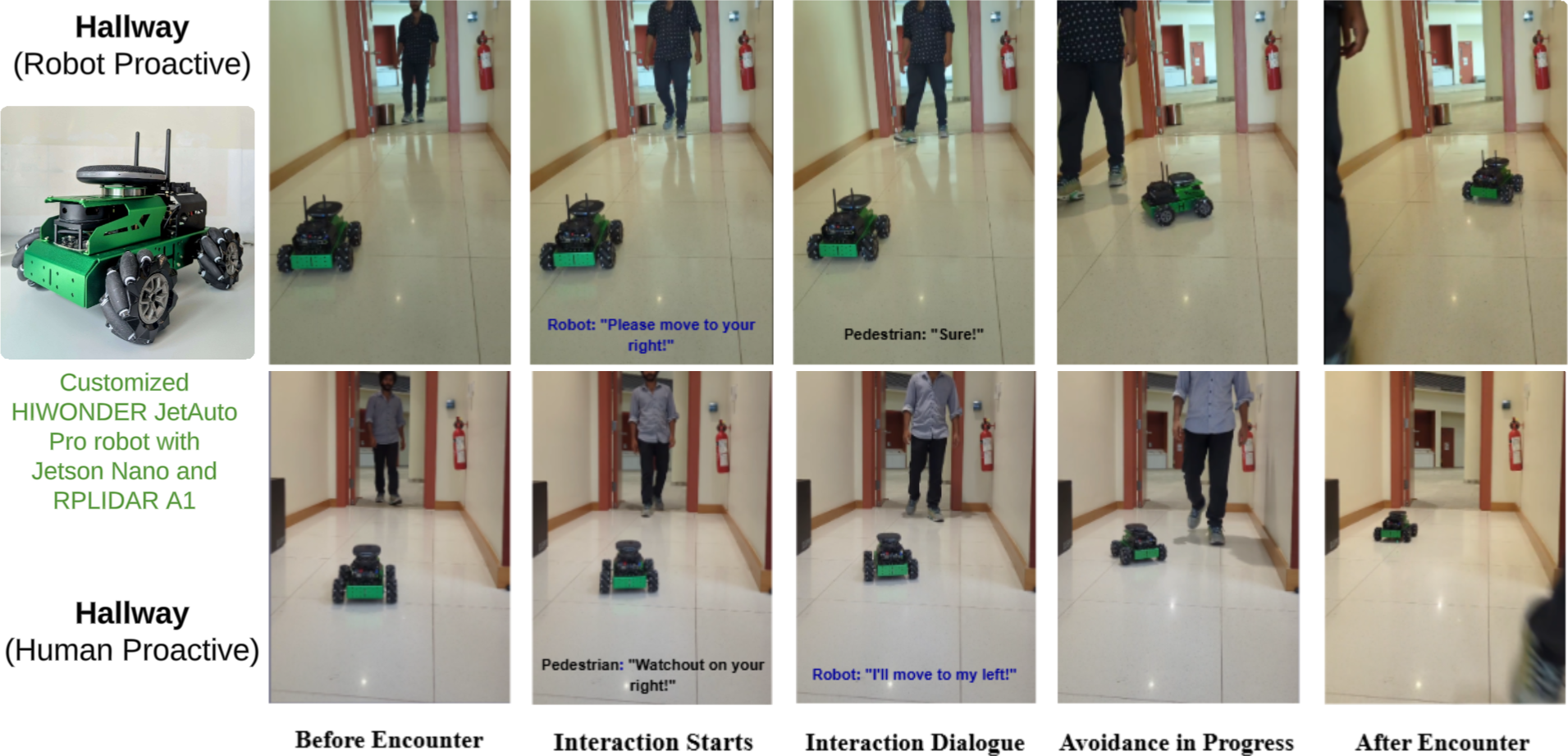}
    \caption{Real-world hallway experiments showing Robot Proactive (top) and Human Proactive (bottom) scenarios, demonstrates how natural-language communication enables effective collision avoidance between robot and pedestrian.}
    \label{fig:real_env}
\end{figure*}

\subsection{Experiments in Real-World environment}

To evaluate our model's practical efficacy, we implemented HSAC-LLM on a customized Hiwonder JetAuto Pro robot, as shown in Fig.~\ref{fig:real_env}. This platform integrates an NVIDIA Jetson Nano (quad-core Cortex-A57 processor at 1.43 GHz with 128 Maxwell CUDA cores delivering 472 GFLOPS), an architecture well-suited for edge-based machine learning inference. The robot is equipped with an RPLIDAR A1 sensor, offering a 5–10 Hz scanning frequency, a 12-meter maximum range, and 1° angular resolution, enabling precise 2D environmental perception without additional visual sensors. The JetAuto Pro's motorized chassis ensures reliable navigation, while the Jetson Nano's energy efficiency supports real-time inference with minimal power consumption.

Our real-world hallway experiments demonstrated two key interaction scenarios shown in Fig.~\ref{fig:real_env}. Each scenario progresses through five phases: Before Encounter, Interaction Starts, Interaction Dialogue, Avoidance in Progress, and After Encounter. In the Robot Proactive scenario (top row), the robot requests "Please move to your right!" and the human complies. In the Human Proactive scenario (bottom row), when warned "Watch out on your right!", the robot responds "I'll move to my left!" and adjusts accordingly. Results confirm that HSAC-LLM achieves robust real-time decision-making and collision avoidance using only a 2D LiDAR. This integration of natural language understanding with navigation validates our model's feasibility for real-world applications requiring effective human-robot communication.


\begin{table}[!h]
\centering
\caption{Performance of HSAC-LLM trained in 2-D and 4-D discrete action space, best values bolded.}
\resizebox{0.48\textwidth}{!}{
\begin{tabular}{  c | c | c| c | c | c }
  \toprule
  \textbf{Scene}             & \textbf{Method} & \textbf{Success}$\uparrow$ & \textbf{Collision}$\downarrow$ & \textbf{Step}$\downarrow$ & \textbf{Training Time}$\downarrow$ \\ [1mm]
  \hline
  \makecell{Square} & 
  \makecell{HSAC-LLM(2-D) \\ HSAC-LLM(4-D)} & 
  \makecell{0.90 $\pm$ 0.029\\ \textbf{0.91 $\pm$ 0.022}} & 
  \makecell{0.10 $\pm$ 0.029\\ \textbf{0.08 $\pm$ 0.022}} & 
  \makecell{20.84 \\ \textbf{19.96}} & 
  \makecell{\textbf{1.45} \\ 1.60} \\ 
  \midrule
  \makecell{Hallway} & 
  \makecell{HSAC-LLM(2-D) \\ HSAC-LLM(4-D)} & 
  \makecell{0.92 $\pm$ 0.036 \\ \textbf{0.93 $\pm$ 0.013} }& 
  \makecell{0.08 $\pm$ 0.036 \\ \textbf{0.07 $\pm$ 0.013} }& 
  \makecell{29.56 \\ \textbf{27.10}} & 
  \makecell{\textbf{1.67} \\ 1.89} \\ 
  \midrule
  \makecell{Crosswalk} & 
  \makecell{HSAC-LLM(2-D) \\ HSAC-LLM(4-D)} & 
  \makecell{0.87 $\pm$ 0.026\\ \textbf{0.91 $\pm$ 0.022} }& 
  \makecell{0.13 $\pm$ 0.026\\ \textbf{0.09 $\pm$ 0.022} }& 
  \makecell{35.22 \\ \textbf{32.80}} & 
  \makecell{\textbf{1.78} \\ 2.04} \\  
  \bottomrule
\end{tabular}}
\label{table:2}
\end{table}

\subsection{Ablation Study}

\subsubsection{Discrete Action Space Dimension}


To assess how discrete action space dimensionality affects navigation performance, we tested HSAC-LLM with both 2-D ($\bm{a}_c \in \{\text{none}, \text{stop}\}$) and 4-D discrete action spaces across Square, Hallway, and Crosswalk environments. Table~\ref{table:2} presents the results based on Success Rate, Collision Rate, Steps Taken, and Training Time metrics. The 4-D discrete action space consistently outperformed the 2-D configuration across all environments. In the Square environment, success rate improved from 0.90 $\pm$ 0.029 to 0.91 $\pm$ 0.022, while collision rate decreased from 0.10 $\pm$ 0.029 to 0.08 $\pm$ 0.022. Similar improvements occurred in the Hallway scenario (success rate: 0.92 $\pm$ 0.036 to 0.93 $\pm$ 0.013; collision rate: 0.08 $\pm$ 0.036 to 0.07 $\pm$ 0.013). The Crosswalk environment showed the most significant improvement, with success rate increasing from 0.87 $\pm$ 0.026 to 0.91 $\pm$ 0.022 and collision rate decreasing from 0.13 $\pm$ 0.026 to 0.09 $\pm$ 0.022. While the 4-D configuration required slightly longer training times (e.g., increasing from 1.67 to 1.89 in Hallway and from 1.78 to 2.04 in Crosswalk), the performance gains justify this trade-off. These results demonstrate that expanding the discrete action space enhances HSAC-LLM's ability to navigate complex environments safely and efficiently, particularly in scenarios requiring nuanced interactions with pedestrians.

\begin{table}[h!]
\centering
\caption{Performance of HSAC-LLM trained in the 1-way and 2-way interaction scenarios, best values bolded.}
\resizebox{0.48\textwidth}{!}{
\begin{tabular}{  c | c | c| c | c | c }
  \toprule
  \textbf{Scene}             & \textbf{Method} & \textbf{Success}$\uparrow$ & \textbf{Collision}$\downarrow$ & \textbf{Step}$\downarrow$ & \textbf{Human Collisions}$\downarrow$\\ 
  \midrule
  \makecell{Square} & 
  \makecell{HSAC \\ HSAC-LLM} & 
  \makecell{{0.86 $\pm$ 0.023} \\ \textbf{0.91 $\pm$ 0.022}} & 
  \makecell{{0.14 $\pm$ 0.023} \\  \textbf{0.08 $\pm$ 0.022}} & 
  \makecell{20.02 \\ \textbf{{19.96}}} & 
  \makecell{0.14 \\ \textbf{0.07}} \\ 
  \midrule
  \makecell{Hallway} & 
  \makecell{HSAC \\ HSAC-LLM} & 
  \makecell{{0.91 $\pm$ 0.018} \\ \textbf{0.93 $\pm$ 0.013} }& 
  \makecell{{0.09 $\pm$ 0.018} \\ \textbf{0.07 $\pm$ 0.013} }& 
  \makecell{29.00 \\ \textbf{27.10}} & 
  \makecell{0.09 \\ \textbf{0.06}} \\ 
  \midrule
  \makecell{Crosswalk} & 
  \makecell{HSAC \\ HSAC-LLM} & 
  \makecell{{0.87 $\pm$ 0.026} \\ \textbf{0.91 $\pm$ 0.022} }& 
  \makecell{{0.13 $\pm$ 0.026} \\ \textbf{0.09 $\pm$ 0.022} }& 
  \makecell{34.78 \\ \textbf{{32.80}}} & 
  \makecell{0.13 \\ \textbf{0.08}} \\  
  \bottomrule
\end{tabular}}
\label{table:3}
\end{table}

\subsubsection{Effectiveness of LLM module}

To evaluate the LLM component's effectiveness in HSAC-LLM, we conducted experiments comparing the full model against a version with the LLM module removed (HSAC). Without the LLM module, the robot could only issue warnings to pedestrians but couldn't process their verbal feedback. Table~\ref{table:3} compares both models across three scenarios. Results show that incorporating the LLM module improved success rates and reduced collision rates and navigation steps in all scenarios. Collision rate improvements were particularly significant in Square and Crosswalk environments, with reductions of 42.86\% and 30.77\%, respectively. The LLM integration enables better natural language understanding and communication with pedestrians, preventing collisions caused by communication gaps. This enhancement not only improves navigational safety but also significantly advances the robot's social navigation capabilities, demonstrating the value of sophisticated language processing in human-robot interactions within dynamic environments.


\section{CONCLUSIONS}

This paper presented HSAC-LLM, a novel framework integrating deep reinforcement learning with large language models for socially-aware robot navigation. Our approach enables natural language interactions between robots and humans, enhancing the robot's ability to understand human intentions in dynamic environments. HSAC-LLM models both continuous navigation actions and discrete conversation actions. Experiments in 2D simulations, Gazebo, and real-world settings demonstrate superior performance in safety, efficiency, and human comfort compared to state-of-the-art DRL methods. This work advances intelligent human-robot interactions in social environments. Future work will focus on improving language understanding, real-time adaptation to complex scenarios, and large-scale real-world deployment.

\addtolength{\textheight}{-3cm}   









\bibliographystyle{IEEEtran}
\bibliography{biblography}

\begin{thebibliography}{10}
\providecommand{\url}[1]{#1}
\csname url@rmstyle\endcsname
\providecommand{\newblock}{\relax}
\providecommand{\bibinfo}[2]{#2}
\providecommand\BIBentrySTDinterwordspacing{\spaceskip=0pt\relax}
\providecommand\BIBentryALTinterwordstretchfactor{4}
\providecommand\BIBentryALTinterwordspacing{\spaceskip=\fontdimen2\font plus
\BIBentryALTinterwordstretchfactor\fontdimen3\font minus \fontdimen4\font\relax}
\providecommand\BIBforeignlanguage[2]{{%
\expandafter\ifx\csname l@#1\endcsname\relax
\typeout{** WARNING: IEEEtran.bst: No hyphenation pattern has been}%
\typeout{** loaded for the language `#1'. Using the pattern for}%
\typeout{** the default language instead.}%
\else
\language=\csname l@#1\endcsname
\fi
#2}}

\bibitem{zhao2024triplemixer}
X.~Zhao, C.~Wen, Y.~Wang, H.~Bai, and W.~Dou, ``Triplemixer: A 3d point cloud denoising model for adverse weather,'' \emph{arXiv preprint arXiv:2408.13802}, 2024.

\bibitem{10.1109/ICRA48506.2021.9561188}
R.~B. Grando, J.~C. de~Jesus, V.~A. Kich, A.~H. Kolling, N.~P. Bortoluzzi, P.~M. Pinheiro, A.~A. Neto, and P.~L.~J. Drews, ``Deep reinforcement learning for mapless navigation of a hybrid aerial underwater vehicle with medium transition,'' in \emph{2021 IEEE International Conference on Robotics and Automation (ICRA)}, 2021, p. 1088–1094.

\bibitem{wen2025zero}
C.~Wen, Y.~Huang, H.~Huang, Y.~Huang, S.~Yuan, Y.~Hao, H.~Lin, Y.-S. Liu, and Y.~Fang, ``Zero-shot object navigation with vision-language models reasoning,'' in \emph{International Conference on Pattern Recognition}.\hskip 1em plus 0.5em minus 0.4em\relax Springer, 2025, pp. 389--404.

\bibitem{wen2024secure}
C.~Wen, J.~Liang, S.~Yuan, H.~Huang, and Y.~Fang, ``How secure are large language models (llms) for navigation in urban environments?'' \emph{arXiv preprint arXiv:2402.09546}, 2024.

\bibitem{kruse2013human}
T.~Kruse, A.~K. Pandey, R.~Alami, and A.~Kirsch, ``Human-aware robot navigation: A survey,'' \emph{Robotics and Autonomous Systems}, vol.~61, no.~12, pp. 1726--1743, 2013.

\bibitem{helbing1995social}
D.~Helbing and P.~Molnar, ``Social force model for pedestrian dynamics,'' \emph{Physical review E}, vol.~51, no.~5, p. 4282, 1995.

\bibitem{7780479}
A.~Alahi, K.~Goel, V.~Ramanathan, A.~Robicquet, L.~Fei-Fei, and S.~Savarese, ``Social lstm: Human trajectory prediction in crowded spaces,'' in \emph{2016 IEEE Conference on Computer Vision and Pattern Recognition (CVPR)}, 2016, pp. 961--971.

\bibitem{gupta2018social}
A.~Gupta, J.~Johnson, L.~Fei-Fei, S.~Savarese, and A.~Alahi, ``Social gan: Socially acceptable trajectories with generative adversarial networks,'' in \emph{Proceedings of the IEEE conference on computer vision and pattern recognition}, 2018, pp. 2255--2264.

\bibitem{keskin2012hand}
C.~Keskin, F.~K{\i}ra{\c{c}}, Y.~E. Kara, and L.~Akarun, ``Hand pose estimation and hand shape classification using multi-layered randomized decision forests,'' in \emph{Computer Vision--ECCV 2012: 12th European Conference on Computer Vision, Florence, Italy, October 7-13, 2012, Proceedings, Part VI 12}.\hskip 1em plus 0.5em minus 0.4em\relax Springer, 2012, pp. 852--863.

\bibitem{9341519}
M.~Nishimura and R.~Yonetani, ``L2b: Learning to balance the safety-efficiency trade-off in interactive crowd-aware robot navigation,'' in \emph{2020 IEEE/RSJ International Conference on Intelligent Robots and Systems (IROS)}, 2020, pp. 11\,004--11\,010.

\bibitem{ma2022reinforcement}
W.~Ma, X.~Gao, J.~Shi, H.~Hu, C.~Wang, Y.~Liang, and O.~Karakus, ``Reinforcement learning based voice interaction to clear path for robots in elevator environment,'' 2022.

\bibitem{dugas2020ian}
D.~Dugas, J.~Nieto, R.~Siegwart, and J.~J. Chung, ``Ian: Multi-behavior navigation planning for robots in real, crowded environments,'' in \emph{2020 IEEE/RSJ International Conference on Intelligent Robots and Systems (IROS)}.\hskip 1em plus 0.5em minus 0.4em\relax IEEE, 2020, pp. 11\,368--11\,375.

\bibitem{yao2021crowd}
S.~Yao, G.~Chen, Q.~Qiu, J.~Ma, X.~Chen, and J.~Ji, ``Crowd-aware robot navigation for pedestrians with multiple collision avoidance strategies via map-based deep reinforcement learning,'' in \emph{2021 IEEE International Conference on Intelligent Robots and Systems}, 2021.

\bibitem{1638022}
H.~Durrant-Whyte and T.~Bailey, ``Simultaneous localization and mapping: part i,'' \emph{IEEE Robotics and Automation Magazine}, vol.~13, no.~2, pp. 99--110, 2006.

\bibitem{temeltas2008slam}
H.~Temeltas and D.~Kayak, ``Slam for robot navigation,'' \emph{IEEE Aerospace and Electronic Systems Magazine}, vol.~23, no.~12, pp. 16--19, 2008.

\bibitem{hart1968formal}
P.~E. Hart, N.~J. Nilsson, and B.~Raphael, ``A formal basis for the heuristic determination of minimum cost paths,'' \emph{IEEE transactions on Systems Science and Cybernetics}, vol.~4, no.~2, pp. 100--107, 1968.

\bibitem{balasuriya2016outdoor}
B.~Balasuriya, B.~Chathuranga, B.~Jayasundara, N.~Napagoda, S.~Kumarawadu, D.~Chandima, and A.~Jayasekara, ``Outdoor robot navigation using gmapping based slam algorithm,'' in \emph{2016 moratuwa engineering research conference (mercon)}.\hskip 1em plus 0.5em minus 0.4em\relax IEEE, 2016.

\bibitem{banino2018vector}
A.~Banino, C.~Barry, B.~Uria, C.~Blundell, T.~Lillicrap, P.~Mirowski, A.~Pritzel, M.~J. Chadwick, T.~Degris, J.~Modayil, \emph{et~al.}, ``Vector-based navigation using grid-like representations in artificial agents,'' \emph{Nature}, vol. 557, no. 7705, pp. 429--433, 2018.

\bibitem{tai2017virtual}
L.~Tai, G.~Paolo, and M.~Liu, ``Virtual-to-real deep reinforcement learning: Continuous control of mobile robots for mapless navigation,'' in \emph{2017 IEEE/RSJ international conference on intelligent robots and systems (IROS)}.\hskip 1em plus 0.5em minus 0.4em\relax IEEE, 2017, pp. 31--36.

\bibitem{zhang2019vr}
J.~Zhang, L.~Tai, P.~Yun, Y.~Xiong, M.~Liu, J.~Boedecker, and W.~Burgard, ``Vr-goggles for robots: Real-to-sim domain adaptation for visual control,'' \emph{IEEE Robotics and Automation Letters}, vol.~4, no.~2, pp. 1148--1155, 2019.

\bibitem{chen2021adoption}
C.~Chen, E.~Demir, Y.~Huang, and R.~Qiu, ``The adoption of self-driving delivery robots in last mile logistics,'' \emph{Transportation research part E: logistics and transportation review}, vol. 146, p. 102214, 2021.

\bibitem{del2019lindsey}
F.~Del~Duchetto, P.~Baxter, and M.~Hanheide, ``Lindsey the tour guide robot-usage patterns in a museum long-term deployment,'' in \emph{2019 28th IEEE international conference on robot and human interactive communication (RO-MAN)}.\hskip 1em plus 0.5em minus 0.4em\relax IEEE, 2019, pp. 1--8.

\bibitem{reis2020service}
J.~Reis, N.~Mel{\~a}o, J.~Salvadorinho, B.~Soares, and A.~Rosete, ``Service robots in the hospitality industry: The case of henn-na hotel, japan,'' \emph{Technology in Society}, vol.~63, p. 101423, 2020.

\bibitem{KRUSE20131726}
\BIBentryALTinterwordspacing
T.~Kruse, A.~K. Pandey, R.~Alami, and A.~Kirsch, ``Human-aware robot navigation: A survey,'' \emph{Robotics and Autonomous Systems}, vol.~61, no.~12, pp. 1726--1743, 2013. [Online]. Available: \url{https://www.sciencedirect.com/science/article/pii/S0921889013001048}
\BIBentrySTDinterwordspacing

\bibitem{he2022multirobot}
Z.~He, C.~Song, and L.~Dong, ``Multi-robot social-aware cooperative planning in pedestrian environments using multi-agent reinforcement learning,'' 2022.

\bibitem{neunert2020continuousdiscrete}
M.~Neunert, A.~Abdolmaleki, M.~Wulfmeier, T.~Lampe, J.~T. Springenberg, R.~Hafner, F.~Romano, J.~Buchli, N.~Heess, and M.~Riedmiller, ``Continuous-discrete reinforcement learning for hybrid control in robotics,'' 2020.

\bibitem{9811662}
Q.~Qiu, S.~Yao, J.~Wang, J.~Ma, G.~Chen, and J.~Ji, ``Learning to socially navigate in pedestrian-rich environments with interaction capacity,'' in \emph{2022 International Conference on Robotics and Automation (ICRA)}, 2022, pp. 279--285.

\bibitem{bester2019multipass}
C.~J. Bester, S.~D. James, and G.~D. Konidaris, ``Multi-pass q-networks for deep reinforcement learning with parameterised action spaces,'' 2019.

\bibitem{delalleau2019discrete}
O.~Delalleau, M.~Peter, E.~Alonso, and A.~Logut, ``Discrete and continuous action representation for practical rl in video games,'' 2019.

\bibitem{chen2020distributed}
G.~Chen, S.~Yao, J.~Ma, L.~Pan, Y.~Chen, P.~Xu, J.~Ji, and X.~Chen, ``Distributed non-communicating multi-robot collision avoidance via map-based deep reinforcement learning,'' \emph{Sensors}, 2020.

\bibitem{4543489}
J.~van~den Berg, M.~Lin, and D.~Manocha, ``Reciprocal velocity obstacles for real-time multi-agent navigation,'' in \emph{2008 IEEE International Conference on Robotics and Automation}, 2008, pp. 1928--1935.

\bibitem{francis2023principles}
A.~Francis, C.~P{\'e}rez-d'Arpino, C.~Li, F.~Xia, A.~Alahi, R.~Alami, A.~Bera, A.~Biswas, J.~Biswas, R.~Chandra, \emph{et~al.}, ``Principles and guidelines for evaluating social robot navigation algorithms,'' \emph{arXiv preprint arXiv:2306.16740}, 2023.

\bibitem{mnih2016asynchronous}
V.~Mnih, A.~P. Badia, M.~Mirza, A.~Graves, T.~P. Lillicrap, T.~Harley, D.~Silver, and K.~Kavukcuoglu, ``Asynchronous methods for deep reinforcement learning,'' 2016.

\bibitem{schulman2017proximal}
J.~Schulman, F.~Wolski, P.~Dhariwal, A.~Radford, and O.~Klimov, ``Proximal policy optimization algorithms,'' 2017.

\bibitem{fujimoto2018addressing}
S.~Fujimoto, H.~van Hoof, and D.~Meger, ``Addressing function approximation error in actor-critic methods,'' 2018.

\bibitem{7139259}
A.~Leigh, J.~Pineau, N.~Olmedo, and H.~Zhang, ``Person tracking and following with 2d laser scanners,'' in \emph{2015 IEEE International Conference on Robotics and Automation (ICRA)}, 2015, pp. 726--733.

\end{thebibliography}

\end{document}